# Overview of FPGA deep learning acceleration based on convolutional neural network

Liusimin

*Abstract*—In recent years, deep learning has become more and more mature, and as a commonly used algorithm in deep learning, convolutional neural networks have been widely used in various visual tasks. In the past, research based on deep learning algorithms mainly relied on hardware such as GPUs and CPUs. However, with the increasing development of FPGAs, both field programmable logic gate arrays, it has become the main implementation hardware platform that combines various neural network deep learning algorithms This article is a review article, which mainly introduces the related theories and algorithms of convolution. It summarizes the application scenarios of several existing FPGA technologies based on convolutional neural networks, and mainly introduces the application of accelerators. At the same time, it summarizes some accelerators' under-utilization of logic resources or under-utilization of memory bandwidth, so that they can't get the best performance.

## I. INTRODUCTION

At present, FPGA technology based on convolutional neural networks is constantly developing and is applied to many practical scenarios. Its performance is constantly improving. While the internal logic resources are continuously increasing and optimized, the internal bandwidth is gradually expanded. However, deep learning algorithms based on convolutional neural networks cannot make full use of these on-chip resources and memory bandwidth, which greatly reduces the utilization rate of the platform and is a waste of resources. At the same time, this is the main problem faced by this technology. The key points that are in urgent need of breakthroughs. This article focuses on the optimization and improvement of FPGA accelerators based on convolutional neural networks, and summarizes the relevant solutions proposed in recent years. These solutions mainly start from the parallelism of FPGA.

First of all, this review article is mainly divided into five parts. The first part is the basic introduction of convolutional neural network and FPGA and related work. The second part mainly introduces the application scenarios of CNN-based FPGA technology. The third part briefly summarizes the basic principles of convolution, and summarizes some methods of accelerator optimization design, the fourth part is the relevant experimental conclusions, and finally the full text summary.

## II. RELATED WORK

### A. Convolutional neural network

Convolutional neural network is a type of feedforward neural network that includes convolution calculations and has a deep structure. It is an important algorithm for deep learning. Its structure includes: input layer, hidden layer, convolution layer, pooling layer, fully connected layer And output layer etc. At present, convolutional neural networks are mainly used in image recognition and processing, video processing fields, such as image feature extraction and abstraction through convolution kernels to achieve image processing, which can greatly reduce the order of magnitude of network training and make neural networks simpler.

### B. FPGA

FPGA appears as a semi-custom circuit in the field of application specific integrated circuits. The basic structure of FPGA includes programmable input and output units, configurable logic blocks, digital clock management modules, embedded block RAM, wiring resources, and embedded dedicated hardware Core, embedded functional unit in the bottom layer. It not only solves the shortcomings of customized circuits, but also overcomes the shortcomings of the limited number of original programmable device circuits.

### C. Combination of CNN and FPGA

The FPGA based on convolutional neural network has its unique advantages in realizing convolutional neural network algorithms by virtue of its high parallel computing, low power consumption, and repeatable configuration.

## III. APPLICATION SCENARIO

In recent years, this technology has also been applied to many scenarios.

### A. accelerator

FPGA parallel accelerator based on convolutional neural network, which is based on the characteristics of convolutional neural network, combined with the pipeline design parallel structure of FPGA to obtain better performance than GPU, CPU and other hardware have faster calculation and processing speed, and realize acceleration function.

## B. Underwater image recognition

FPGA embedded real-time image recognition based on convolutional neural network also takes advantage of FPGA's low power consumption and strong computing power, and uses its parallel technology Multi-depth parallelization in convolution operation is realized. The accuracy difference between the obtained FPGA system and the ground workstation is very slight. It can be approximately considered that the same accuracy can be achieved, which solves the problem of difficult underwater recognition;

## C. Speech Recognition

In addition to being applied to common image and accelerator fields, CNN-based FPGAs are also used Applied to speech recognition, with the continuous development of intelligence, the technical requirements for image recognition and speech recognition are also constantly improving. FPGA implementation based on convolution algorithms can better continuously improve performance to meet such needs, such as using maximum The pooled convolution model greatly reduces the volume of the entire convolution model and the amount of related calculations. In the implementation of FPGA, 16-bit fixed-point numbers are used to replace floating-point numbers for operation while ensuring accuracy and reducing FPGA storage At the same time, a parallel structure is adopted to accelerate the realization of operations. This design greatly achieves the accuracy and speed of speech recognition. This is a feature of CNN-based FPGA technology that is significantly better than other algorithms or hardware platforms.

## IV. IMPROVEMENT AND OPTIMIZATION TECHNOLOGY

### A. Convolution principle

The convolutional layer of the convolutional neural network is composed of several feature maps, and each feature map is composed of several neurons, and these neurons represent the image extracted from the features. These convolution images share the same weight value through the same convolution kernel convolution, so the weight value sharing mechanism greatly reduces the number of training sets required by the neural network. If the convolutional layer extracts N feature maps, they are convolved by the same K*k convolution kernel to generate M feature maps for the next layer of convolutional layer.

The excitation layer is to do a nonlinear mapping to the output of the convolutional layer. Commonly used excitation functions are: sigmoid function, Tanh function, ReLU function, LeakyReLU function, ELU function, Maxout function,

Pooling layer: Pooling, also known as "undersampling" or "downsampling", is mainly used to reduce dimensionality and compress the number of data and parameters, reduce overfitting and improve fault tolerance.

Output layer: Also known as the "fully connected layer", the feature map after convolution, excitation, and pooling is fully connected.

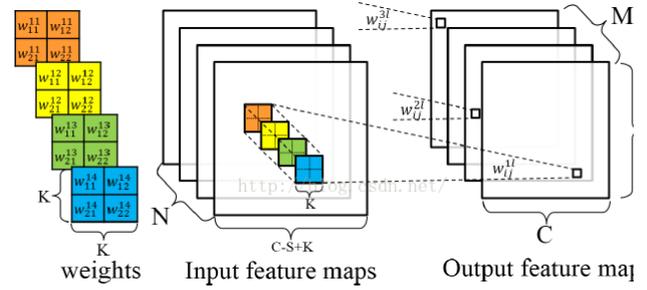

Fig. 1. Gragh of a convolutional layer (The picture is taken from the Internet)

### B. Optimization and improvement

*1)*

From the basic concept of convolutional neural network: the framework of the convolution module is mainly control module, data input module, calculation module, data output module, and the calculation module is composed of multiple sets of vector units, and the length of the calculation unit is floating point For convolution calculations, the floating-point addition unit is relatively cumbersome and cannot achieve high performance. At present, one solution is to use a pipeline floating-point adder. Because the output of the adder needs to go through the accumulation process, and this process needs to be driven by the clock, and because the output of each layer is passed from one level to the next, the input of the next layer depends on the output of the previous layer, so it needs There is control logic to ensure that the calculation process of the convolution is at the correct time node.

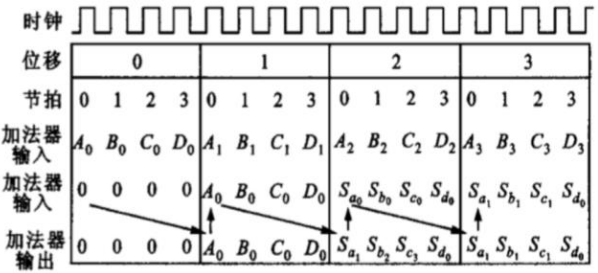

Fig. 2. Schematic diagram of floating-point pipeline filling (Excerpt from Literature 7)

This method also addresses the problem of insufficient efficiency caused by the existence of empty beats in the pipeline. It proposes that the control logic unit on the beat of the pipeline is inserted into the convolution kernels belonging to different output channels for calculation, so as to make the processor internal The operation between the modules is more efficient, achieving the goal of making full use of the resources inside the hardware and reducing power consumption.At present, many optimization measures are based on parallelism and pipeline. For example, other feasible optimization schemes proposed by scholars: there are often many highly repetitive data in the calculation process, and FPGA is parallel, which requires More computing parallelism, which will also reduce data utilization. The image information acquired in the internal memory is stored in REG, and the weight of the convolution kernel in the memory is stored in W. The modification operation can only be performed when the convolution operation needs to update the convolution kernel.

*2)*

The CNN convolution algorithm generally involves multiplication and addition. Some scholars have analyzed the parameters involved in the calculation of the convolution kernel. The process is to first ignore the 0 value of the weight in the convolution kernel, only consider the 0 data of the input feature map, and find that it has sparseness. This is of great help to the simplification of the convolution calculation, although this method is used in many models The improvement effect is not obvious, but when used in FPGA implementation, the simplification benefits brought by sparsity can be used effectively. Although the scholar also mentioned that adders and comparators are added in the process, it is compared with multiplication. The feasibility of this method is obvious for the number and complexity of the processor, which significantly simplifies the calculation process and reduces the calculation speed.

The parallel rectangular multiplier is used for calculation, and 0 data is directly input from A, which can omit many unnecessary calculations. Its structure is shown in the figure3.

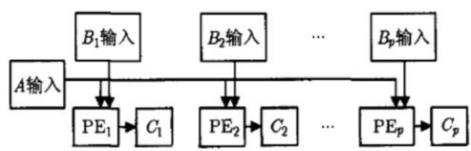

Fig. 3. Schematic diagram of matrix multiplier structure (Excerpt from Literature 5)

*3)*

According to the characteristics of the time-frequency representation of human mD signals, some scholars have proposed an FPGA-based CNN acceleration classification method based on millimeter wave mD signals. The calculation and operation process are optimized by combining the operation of CNN and the structural characteristics of FPGA, which improves the execution speed of FPGA and greatly reduces the waiting time and cache time.

The main idea is: First, use the high Doppler sensitivity of millimeter-wave radar, use the mD spectrogram represented in the time-frequency domain as the CNN input, and secondly perform a series of quantization operations on the different data types of the CNN model: first normalize the spectrogram Converted to the interval [0,1], the weights are quantified, and the feature map is dynamically quantized in consideration of the changes in the data between layers; thirdly, when the intra-channel convolution is implemented, not only the calculation parallel scheme is used, but also the data Two schemes combined with it in parallel. Data is passed from the previous layer to the current convolutional layer in parallel, which is data parallel. Then along the depth direction, all the two-dimensional data matrices in the dimension are convolved with the corresponding kernels simultaneously, and the calculation is parallel. Finally, the matrix of these results is added in parallel with the offset parameters to obtain the output of the channel. This effectively improves the problem that a large number of convolutional layers consume resources.

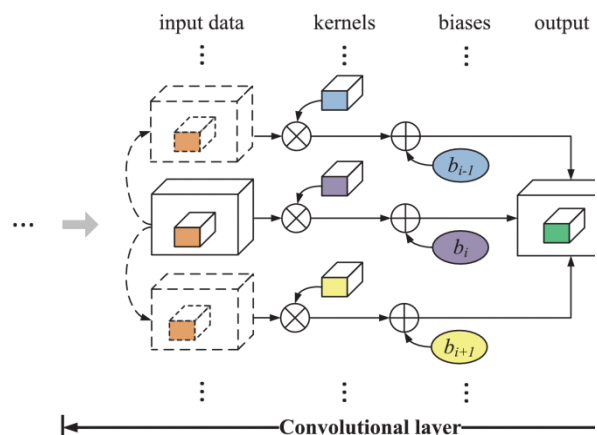

Fig. 4. Schematic diagram of parallel calculation of the middle channel of the convolutional layer.(Excerpt from Literature 2)

Finally, hardware experiments were carried out, using radar to collect a real-world radar data set of real human targets, and at the same time a series of index calculations were carried out on FPGA. The results showed that the proposed FPGA-based CNN acceleration classification based on millimeter wave mD signals In practical applications, the method can increase the speed while ensuring the accuracy.

## V. EXPERIMENT

*A.*

The first optimization method mentioned above is to use a pipelined floating-point adder to improve accelerator performance. The scholars of this study used the VC707 suite of XILINX as the hardware test platform. The first is to test on different platforms and get the conclusion: Using this method, the FPGA-based convolutional neural network platform has less running time and better performance. Then compare other mainstream accelerator design methods in this period to get the calculation rate shown in Table 2, which shows that this research method is of great significance under the condition of single precision.

*B.*

The matrix multiplier designed in this research is implemented on VIVADO HLS. At the same time, in order to

TABLE I
TIME-CONSUMING RATIO OF DIFFERENT PLATFORMS
(Excerpt from Literature 7)

| Test set | operation hours E5-2640 V4 | operation hours GTX-1080 | operation hours FPGA | Relative running time ratio |
|---|---|---|---|---|
| MNIST-LeNet | 0.231 | 0.0032 | 0.086 | 1:72:2.7 |
| *CIFAR-10* | 1.234 | *0.0108* | 0.792 | 1:114:1.6 |
| AlexNet | 72.843 | *0.2940* | 39.07 | 1:248:1.9 |

TABLE II
PERFORMANCE COMPARISON WITH OTHER DOCUMENTS
(Excerpt from Literature 7)

| Program characteristics | Device model | Clock frequency | Numerical format | calculation rate |
|---|---|---|---|---|
| This design | Virtex-7 485T | 100 | Single precision | 55.11 |
| *Literature [16]* | Virtex-7 485T | *100* | Single precision | 62.62 |
| Literature [17] | Stratix-VGXA7 | *181* | Single precision | 50.07 |
| Literature [18] | ZynqXC7Z045 | 180 | 16-bit floating point | 187.80 |

compare with other previous research projects, the same platform VIRTEX-7 VC707 EVALUATION PLATFO is selected for verification test, and the calculation time is measured. With the data shown in the table, except for the first data, it can be seen that the time has been reduced compared with the comparative research project. So using sparsity to reduce calculation time is of practical significance

.TABLE III
CALCULATION TIME COMPARIS
(EXCERPT FROM LITERATURE 5)

| | FPGA2015 (32-bit floating point)(ms) | This article realizes (8-bit fixed point) |
|---|---|---|
| Conv1 | 7.67 | 8.02 |
| Conv2 | 5.35 | 5.14 |
| Conv3 | 3.79 | 2.13 |
| *Conv4* | 2.88 | 1.31 |
| Conv5 | 1.93 | 0.88 |
| Total | 21.61 | 17.48 |

*C.*

For the design of the FPGA-based CNN accelerator based on millimeter-wave mD signals mentioned above, the student used a variety of quantization methods to conduct experiments to obtain the memory requirements of CNN before and after quantization under different data types as shown in the tableIV. Seeing that the memory resource consumption has been greatly reduced

TABLE IV
Data Quantization Used for FPGA Based CNN Accelerator
(Excerpt from Literature 2)

| Layers | data | bitwidth | FRACTION | INTEGER BITS | Signed/unsigned |
|---|---|---|---|---|---|
| Conv1 | Input | 16 | 15 | 1 | Unsigned |
| | *Bais* | *32* | 30 | 1 | signed |
| *Conv2* | Input | *16* | 13 | 3 | Unsigned |
| | Bias | *32* | 28 | 3 | signed |
| *Conv2* | Input | *16* | 9 | 7 | Unsigned |
| | Bias | 32 | 24 | 7 | signed |
| FC1 | Input | 16 | 10 | 6 | Unsigned |
| | | *32* | 25 | 6 | signed |
| *FC2* | Input | *16* | 11 | 4 | signed |
| | | *32* | 26 | 5 | signed |
| *Conv1~3 FC1~2* | *Weights* | *16* | 15 | 0 | signed |

VI. CONCLUSION

The above is the recent CNN-based FPGA design optimization method, which mainly uses the optimization pipeline method to optimize the FPGA. As for convolutional networks, many addition and multiplication operations are involved in the convolution operation of the convolutional network. At present, many scholars are starting from this to carry out related optimization studies, thereby simplifying calculations and improving the overall operation speed. The various parameters obtained from actual hardware experiments based on its optimization theory can also prove that these theories are feasible, and they all improve the accelerator performance to a certain extent. However, we can see that most of the operations in the design process use floating-point numbers, but floating-point numbers also have the problem of occupying unnecessary resources. Using fixed-point numbers can relatively solve this problem. This is also mentioned in the previous speech recognition application. Perhaps this is also a strategy for optimizing FPGA accelerators based on convolutional neural networks. From this perspective, you can conduct related research and conduct related experiments based on theories, and compare the two processes. It is hoped that subsequent research can solve the related problems. problem.

This article is the final course assessment of the artificial intelligence overview course and the assessment homework for Mr. Zhou Chuanxi. It aims to let students who study this course understand more about artificial intelligence.